%% file: paper.tex
\documentclass[letterpaper, 10 pt, conference]{ieeeconf}  

\IEEEoverridecommandlockouts                              

\usepackage[printonlyused,nohyperlinks]{acronym}
\usepackage{subcaption}

\usepackage{amssymb}
\usepackage{amsmath}
\usepackage{xspace}
\usepackage{hyperref}
\usepackage{booktabs}
\usepackage{float}

\usepackage{graphicx}
\usepackage{algpseudocode}
\usepackage{algorithm}
\usepackage{multirow}
\usepackage{verbatim}
\usepackage{array}

\usepackage{tikz}
\usetikzlibrary{calc}
\usetikzlibrary{decorations.pathmorphing}
\usepackage{caption}
\usepackage{siunitx}
\input{tex/paper_newcommands}

\newcommand{\citep}[1]{\cite{#1}}

\title{Model-Based Action Exploration for Learning Dynamic Motion Skills}

\author{
Glen Berseth and Michiel van de Panne \\
University of British Columbia
}

\begin{document}

\maketitle

\input{tex/paper_glossary}	

\input{tex/paper_abstract}
\input{tex/paper_Introduction}

\input{tex/paper_related_work}

\input{tex/paper_Framework}

\input{tex/paper_MBAE}

\input{tex/paper_Results}

\input{tex/paper_Discussion}


\bibliographystyle{IEEEtran}
\bibliography{paper}

\input{tex/paper_appendix}

\end{document}

%% file: tex/paper_newcommands.tex
\newcommand{\todo}[1]{\textcolor{red}{Todo:#1}}
\newcommand{\glen}[1]{}

\newcommand{\refequation}[1]{Eq. \ref{#1}}
\newcommand{\refsection}[1]{Section: \ref{#1}}
\newcommand{\refFigure}[1]{Figure~\ref{#1}}

\newcommand{\refAlgorithm}[1]{Algorithm~\ref{#1}}

\newcommand{\function}[1]{\ensuremath{\textit{#1}}}

\newcommand{\normalDistribution}[2]{\mathcal{N}(#1,#2)}
\newcommand{\expectation}{\mathbb{E}}
\newcommand{\lossFunction}[1]{L(#1)}
\newcommand{\grad}[0]{\nabla}

\newcommand{\forwardDyanmicsText}{transition probability\xspace}
\newcommand{\MBAEText}{Model-Based Action Exploration\xspace}

\newcommand{\DDPG}{\ac{DDPG}\xspace}
\newcommand{\CACLA}{\ac{CACLA}\xspace}
\newcommand{\MDP}{\ac{MDP}\xspace}

\newcommand{\ReLU}{\ac{ReLU}\xspace}
\newcommand{\DYNA}{\ac{DYNA}\xspace}
\newcommand{\PPO}{\ac{PPO}\xspace}
\newcommand{\EPG}{\ac{EPG}\xspace}
\newcommand{\DPG}{\ac{DPG}\xspace}

\newcommand{\RL}{\ac{RL}\xspace}
\newcommand{\SVG}{\ac{SVG}\xspace}
\newcommand{\GAN}{\ac{GAN}\xspace}
\newcommand{\cGAN}{\ac{cGAN}\xspace}
\newcommand{\MSE}{\ac{MSE}\xspace}
\newcommand{\Membrane}{\emph{Membrane}\xspace}

\newcommand{\MBAE}{\SMBAE}
\newcommand{\SMBAE}{\ac{MBAE}\xspace}

\newcommand{\stateSpace}{S}
\newcommand{\actionSpace}{A}

\newcommand{\myState}{\sstate}
\newcommand{\sstate}{x}
\newcommand{\observation}{\sstate}
\newcommand{\action}{u}
\newcommand{\reward}{r}
\newcommand{\ttime}{t}

\newcommand{\discountFactor}{\gamma}
\newcommand{\expNoise}{\lambda}
\newcommand{\experianceTuple}{\tau}
\newcommand{\experianceMemory}{D}
\newcommand{\learningRate}{\alpha}

\newcommand{\agent}{agent\xspace}
\newcommand{\character}{2D Biped\xspace}
\newcommand{\reacher}{2D Reacher\xspace}
\newcommand{\halCheetah}{HalfCheetah\xspace}
\newcommand{\policySymbol}{\pi}
\newcommand{\noiseParam}{\eta}

\newcommand{\policy}[1]{\policySymbol(#1)}
\newcommand{\transitionFunction}[1]{P(#1)}
\newcommand{\modelFunction}[1]{p(#1)}
\newcommand{\forwardDynamicsSymbol}{\hat{P}}
\newcommand{\forwardDynamicsMeanSymbol}{\hat{P}_{\mu}}
\newcommand{\forwardDynamicsSTDSymbol}{\hat{P}_{\sigma}}
\newcommand{\forwardDynamics}[1]{\forwardDynamicsSymbol(#1)}
\newcommand{\forwardDynamicsMEAN}[1]{\forwardDynamicsMeanSymbol(#1)}
\newcommand{\forwardDynamicsSTD}[1]{\forwardDynamicsSTDSymbol(#1)}

\newcommand{\policyMean}[1]{\mu(#1)}
\newcommand{\policyVariance}[0]{\Sigma}
\newcommand{\policySTD}[1]{\sigma(#1)}
\newcommand{\valueFunction}[1]{V_{\policySymbol}(#1)}
\newcommand{\actionvalueFunction}[1]{Q_{\policySymbol}(#1)}
\newcommand{\AdvantageSymbol}{A}
\newcommand{\advantageFunction}[1]{\AdvantageSymbol_{\policySymbol}(#1)}
\newcommand{\rewardFunction}[1]{R(#1)}
\newcommand{\modelParametersPolicy}[0]{\theta_{\mu}}
\newcommand{\modelParametersPolicySTD}[0]{\theta_{\sigma}}
\newcommand{\modelParametersStocasticPolicy}[0]{\theta_{\policySymbol}}

\newcommand{\modelParametersValueFunction}[0]{\theta_{v}}
\newcommand{\modelParametersForwardDynamics}[0]{\theta_{\forwardDynamicsSymbol}}
\newcommand{\modelParametersForwardDynamicsMEAN}[0]{\theta_{\forwardDynamicsMeanSymbol}}
\newcommand{\modelParametersForwardDynamicsSTD}[0]{\theta_{\forwardDynamicsSTDSymbol}}

%% file: tex/paper_glossary.tex
\acrodef{UI}{user interface}
\acrodef{UBC}{University of British Columbia}
\acrodef{MDP}{Markov Dynamic Processes}


\acrodef{ANOVA}[ANOVA]{Analysis of Variance\acroextra{, a set of
  statistical techniques to identify sources of variability between groups}}
\acrodef{API}{application programming interface}
\acrodef{CTAN}{\acroextra{The }Common \TeX\ Archive Network}
\acrodef{DOI}{Document Object Identifier\acroextra{ (see
    \url{http://doi.org})}}
\acrodef{GPS}[GPS]{Graduate and Postdoctoral Studies}
\acrodef{PDF}{Portable Document Format}
\acrodef{RCS}[RCS]{Revision control system\acroextra{, a software
    tool for tracking changes to a set of files}}
\acrodef{TLX}[TLX]{Task Load Index\acroextra{, an instrument for gauging
  the subjective mental workload experienced by a human in performing
  a task}}
\acrodef{UML}{Unified Modelling Language\acroextra{, a visual language
    for modelling the structure of software artefacts}}
\acrodef{URL}{Unique Resource Locator\acroextra{, used to describe a
    means for obtaining some resource on the world wide web}}
\acrodef{W3C}[W3C]{\acroextra{the }World Wide Web Consortium\acroextra{,
    the standards body for web technologies}}
\acrodef{XML}{Extensible Markup Language}
\acrodef{MBAE}{Model-Based Action Exploration}
\acrodef{SMBAE}{Stochastic Model-Based Action Exploration}
\acrodef{DRL}{Deep Reinforcement Learning}
\acrodef{HRL}{Hierarchical Reinforcement Learning}
\acrodef{DDPG}{Deep Deterministic Policy Gradient}
\acrodef{CACLA}{Continuous Actor Critic Learning Automaton}
\acrodef{HLC}{High-Level Controller}
\acrodef{LLC}{Low-Level Controller}
\acrodef{ReLU}{Rectified Linear Unit}
\acrodef{PPO}{Proximal Policy Optimization}
\acrodef{EPG}{Expected Policy Gradient}
\acrodef{DPG}{Deterministic Policy Gradient}
\acrodef{DYNA}{DYNA}
\acrodef{GAE}{Generalized Advantage Estimation}
\acrodef{RL}{Rienforcement Learning}
\acrodef{SVG}{Stochastic Value Gradients}
\acrodef{GAN}{Generative Advasarial Network}
\acrodef{cGAN}{Conditional Generative Advasarial Network}
\acrodef{MSE}{Mean Squared Error}

%

%% file: tex/paper_abstract.tex
\begin{abstract}
	
Deep reinforcement learning has achieved great strides in solving challenging motion control tasks.
Recently, there has been significant work on methods for exploiting the data gathered during
training, but there has been less work on how to best generate the data to learn from.  For continuous action
domains, the most common method for generating exploratory actions involves sampling from a Gaussian
distribution centred around the mean action output by a policy.  Although these methods 
can be quite capable, they do not scale well with the dimensionality of the action space, 
and can be dangerous to apply on hardware.
We consider learning a forward dynamics model to predict the result, 
($\myState_{\ttime+1}$), of taking a particular action, ($\action$), given a
specific observation of the state, ($\myState_{\ttime}$).  With this model 
we perform internal look-ahead predictions of outcomes and seek actions we believe 
have a reasonable chance of success.
This method alters the exploratory action space, thereby increasing learning speed and enables higher
quality solutions to difficult problems, such as robotic locomotion and juggling.

\end{abstract}

%% file: tex/paper_Introduction.tex
\label{sec:Intro}


Efficient action exploration remains a challenge for \RL, especially in continuous action domains.
Gaussian distributions are commonly used to model the stochastic component of 
control policies in continuous action spaces.  
However, as the number of action dimensions increase the probability of randomly generating an action that is
better than the current policy decreases exponentially with the action dimension.

Recent work has indicated that methods that add local noise might not be enough to explore
environments well~\citep{2017arXiv170307608O}. Other methods model the action distribution
better by learning a noise distribution~\citep{2017arXiv170610295F} or processing random
Gaussian noise through the policy~\citep{2017arXiv170601905P}.  However, these methods may
not scale well with respect to the number of action dimensions.  In practise, we expect that
there likely exists better distributions that will focus the sampled actions to areas of the
action space that appear more promising.  This is of particular importance to problems where
data collection is expensive, like robotics.
	
We want to generate exploratory actions that have a greater probability of leading the policy to
higher value states.  We do this in a model-based way, using a learned model of the environment's
transition probability $\forwardDynamics{\observation_{\ttime+1}
| \observation_{\ttime}, \action_{\ttime}, \modelParametersForwardDynamics }$.  The model is used to
predict outcomes of taking actions in particular states.  Predicted states are rated, via the value
function, for how well it is believed the policy will perform from that state onward.  This method
is similar to how a person might use an internal understanding of the result of taking an action and
modifying that action to increase the utility (or cumulative reward) of their future.  We call this
method \MBAE.  This work is a step towards mixing model-based and model-free learning.  We
use a model-based method to assist action exploration and model-free methods for training.
We note that \MBAE should be used in conjunction with off-policy algorithms.
	
Our work is motivated by the desire to solve more complex simulated and robotic tasks.  The
mathematical framework is inspired by \DPG~\citep{DBLP:journals/corr/LillicrapHPHETS15}
where action gradients can be computed for the policy given an action-value (Q) function
However, in practise these gradients are noisy and can vary greatly in their
magnitude.  This can make it challenging to create a stable learning framework.  In a sense,
in \MBAE we are passing these gradients through the environment as an extra means of validation,
thereby decreasing the bias and increasing the stability of learning.  This work is also motivated
by the idea that the significant body of data collected while training an \RL policy should be
leveraged for additional purposes.
	
\begin{figure}%
	\centering
	\setlength{\fboxsep}{1pt}%
	\subcaptionbox{\label{fig:Membrane-hardware}Membrane Hardware}{\includegraphics[width=0.45\linewidth]{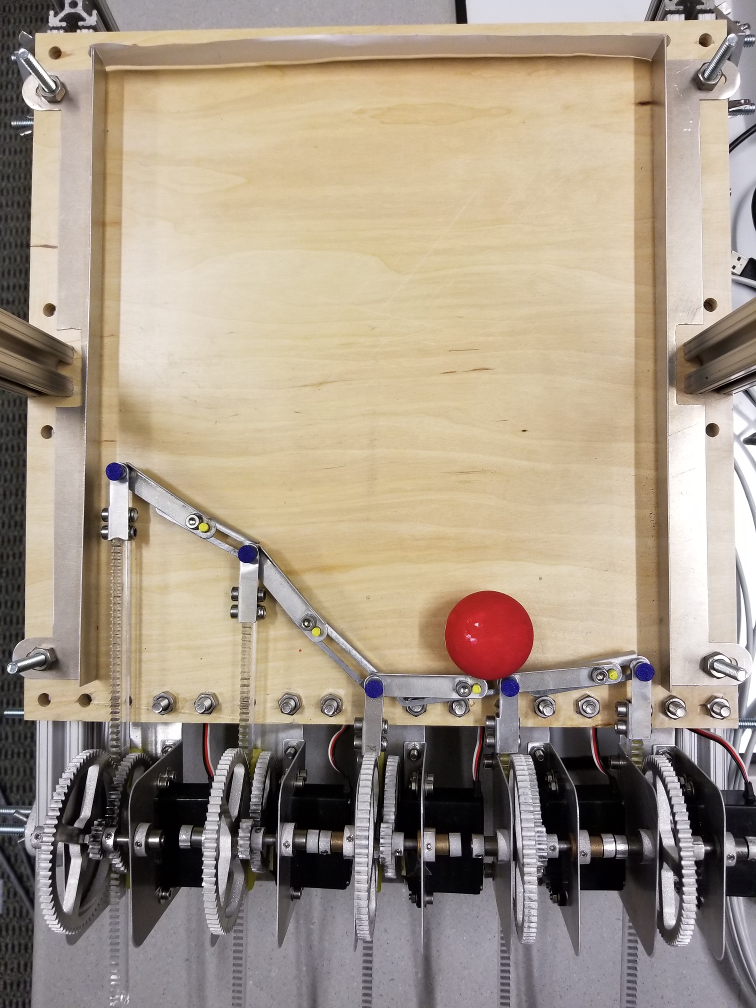}}
	\subcaptionbox{\label{fig:Membrane-software}Membrane Simulation}{\includegraphics[width=0.45\linewidth]{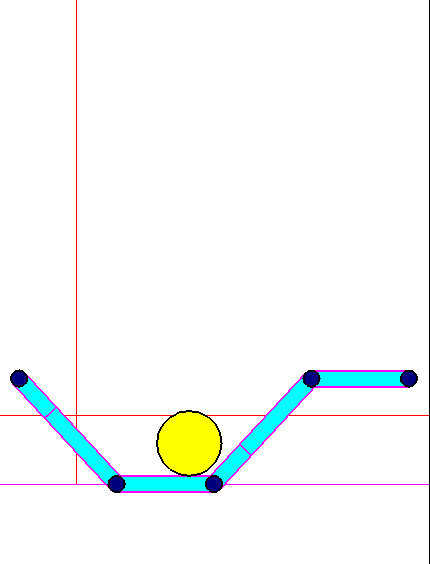}}
	\caption{(a) The \Membrane robot. 
	The blue points between the parallel links are actuated up and down via the servos on the bottom.
	Each point is connected together with a passive slider, together they form a \Membrane-like system.
	Right: Simulated model of \Membrane robot.}
	\label{fig:cassie}
\end{figure}

%% file: tex/paper_related_work.tex
\section{Related Work}
\label{sec:related_work}

\paragraph{Reinforcement Learning}
The \emph{environment} in a RL problem is often modelled as an \MDP with a discrete set of states
and actions~\citep{sutton1998reinforcement}.  In this work we are focusing on problems with
infinite/continuous state and action spaces.  These include complex motor control tasks that have
become a popular benchmark in the machine learning literature~\citep{2017arXiv170702286H}.  Many
recent \RL approaches are based on policy gradient methods~\citep{policygradient} where the gradient
of the policy with respect to future discounted reward is approximated and used to update the
policy.  Recent advances in combining policy gradient methods and deep learning have led to
impressive results for numerous problems, including Atari games and bipedal motion
control~\citep{DBLP:journals/corr/SchulmanMLJA15,A3C,mnih2015human,vanHasselt2012,HeessWTLRS16,
  Peng:2017:DDL:3072959.3073602}.

\paragraph{Sample Efficient RL}
While policy gradient methods provide a general framework for how to update a policy given data, it
is still a challenge to generate good data.  Sample efficient RL methods are an important area of
research as learning complex policies for motion control can take days and physically simulating on
robots is time-consuming.  Learning can be made more sample efficient by further parameterizing the
policy and passing noise through the network as an alternative to adding vanilla Gaussian
noise~\citep{2017arXiv170601905P,2017arXiv170610295F}.  Other work encourages exploration of the
state space that has not yet been seen by the agent~\citep{DBLP:journals/corr/HouthooftCDSTA16}.
There has been success in incorporating model-based methods to generate synthetic data or locally
approximate the
dynamics~\citep{Gu:2016:CDQ:3045390.3045688,DBLP:journals/corr/MishraAM17,sutton1991dyna}.  Two
methods are similar to the \MBAE work that we propose.  Deep Deterministic Policy Gradient (\DDPG)
is a method that directly links the policy and value function, propagating gradients into the policy
from the value function~\citep{DBLP:journals/corr/LillicrapHPHETS15}.  Another is \SVG, a framework
for blending between model-based and model-free learning~\citep{NIPS2015_5796}.  However, these
methods do not use the gradients as a method for action exploration.
	
\paragraph{Model-Based \RL}
generally refers to methods that use the structure of the problem to assist learning.  Typically any
method that uses more than a policy and value function is considered to fall into this category.
Significant improvements have been made recently by including some model-based knowledge into the \RL
problem.  By first learning a policy using model-based RL and then training a model-free method to
act like the model-based method~\citep{2017arXiv170802596N} significant improvements are achieved.
There is also interest in learning and using models of the transition dynamics to improve
learning~\citep{2017arXiv170903153B}.  The work in~\citep{DBLP:journals/corr/MishraAM17} uses
model-based policy optimization methods along with very accurate dynamics models to learn good
policies.  In this work, we learn a model of the dynamics to compute gradients to maximize future
discounted reward for action exploration.  The dynamics model used in this work does not need to be
particularly accurate as the underlying model-free RL algorithm can cope with a noisy action
distribution.


%% file: tex/paper_Framework.tex
\section{Framework}
\label{sec:framework}

In this section we outline the \MDP based framework used to describe the \RL problem.

\subsection{Markov Dynamic Process}


An \MDP is a tuple consisting of $\{\stateSpace, \actionSpace, \rewardFunction{\cdot},
\transitionFunction{\cdot}, \discountFactor\}$.  Here $\stateSpace$ is the space of all possible
state configurations and $\actionSpace$ is the set of available actions.  The reward function
$\rewardFunction{\action, \sstate}$ determines the reward for taking action $\action
\in \actionSpace$ in state $\sstate \in \stateSpace$.  
The probability of ending up in state
$\observation_{\ttime+1} \in \stateSpace$ after taking action $\action$ in state $\sstate$ is
described by the transition dynamics function $\transitionFunction{\observation_{\ttime+1} |
  \sstate, \action }$.  Lastly, the discount factor $\discountFactor \in (0,1]$ controls the
  \emph{planning horizon} and gives preference to more immediate rewards.  A stochastic policy
  $\policy{\action|\sstate}$ models the probability of choosing action $\action$ given state
  $\sstate$. The quality of the policy can be computed as the expectation over future discounted
  rewards for the given policy starting in state $\sstate_{0}$ and taking action $\action_{0}$.

\begin{equation}
\label{eqn:policyQualityFunction}
\begin{split}
J_{\policySymbol}(\sstate_{0}, \action_{0}) & = \mathop{\expectation_{\policySymbol}}[
\rewardFunction{\action_{0}, \sstate_{0}} +
\discountFactor \rewardFunction{\action_{1}, \sstate_{1}} + 
\ldots + 
 \discountFactor^{T} \rewardFunction{\action_{T}, \sstate_{T}}]
\\
J_{\policySymbol}(\sstate_{0}, \action_{0}) & = \mathop{\expectation_{\policySymbol}}[\sum_{\ttime=0}^{T} \discountFactor^{\ttime} \rewardFunction{\action_{\ttime}, \sstate_{\ttime}}]
\end{split}
\end{equation}

The actions $\action_{\ttime}$ over the trajectory $(\action_{0}, \sstate_{0}, \ldots , \action_{T},
\sstate_{T})$ are determined by the policy $\policy{\action_{\ttime}, \sstate_{\ttime}}$.  The
successor state $\observation_{\ttime+1}$ is determined by the transition function
$\transitionFunction{\observation_{\ttime+1} | \sstate_{\ttime}, \action_{\ttime} }$.

\subsection{Policy Learning}

The state-value function $\valueFunction{\sstate}$ estimates \refequation{eqn:policyQualityFunction}
starting from state $\sstate_{0}$ for the policy $\policy{\cdot}$.  The action-valued function
$\actionvalueFunction{\sstate, \action}$ models the future discounted reward for taking action
$\action$ in state $\sstate$ and following policy $\policy{\cdot}$ thereafter.  The
advantage function is a measure of the benefit of taking action $\action$ in state $\sstate$ with
respect to the current policy performance.
\begin{equation}
\label{eqn:advantageFunction}
\advantageFunction{\sstate, \action} = \actionvalueFunction{\sstate, \action} - \valueFunction{\sstate}
\end{equation}
The advantage function is then used as a metric for improving the policy.



\begin{equation}
\label{eqn:policyImprovement}
\action^* = \arg\max \log \policy{\action | \sstate} \advantageFunction{\sstate, \action}
\end{equation}

\subsection{Deep Reinforcement Learning}

During each episode of interaction with the environment, data is collected 
for each action taken, as an \emph{experience} tuple $\experianceTuple_{i}= (\observation_{i}, \action_{i}, \reward_{i},
\observation'_i)$.


\subsubsection{Exploration}

In continuous spaces the stochastic policy $\policy{\action | \sstate}$ is often modeled by a Gaussian
distribution with mean $\policyMean{\sstate | \modelParametersPolicy}$.  The standard deviation can
be modeled by a state-dependent neural network model, $\policySTD{\sstate | \modelParametersPolicySTD}$, or can be
state independent and sampled from $\normalDistribution{0}{\policyVariance}$. 

\subsubsection{Exploitation}

We train a neural network to model the value function on data collected from the policy.  The loss
function used to train the value function ($\valueFunction{\sstate| \modelParametersValueFunction}$)
is the temporal difference error:
\begin{equation}
\label{eqn:temporalDifference}
L(\modelParametersValueFunction) = E [r + \discountFactor \valueFunction{\observation_{\ttime+1} | \modelParametersValueFunction}] - \valueFunction{\sstate | \modelParametersValueFunction}].
\end{equation}
Using the learned value function as a baseline, the advantage function can be estimated from data.
With an estimate of the policy gradient, via the advantage, policy updates can be performed to
increase the policy's likelihood of selecting actions with higher advantage:
\begin{equation}
\label{eqn:policyImprovement2}
 \modelParametersStocasticPolicy \leftarrow \modelParametersStocasticPolicy + \alpha \grad_{\modelParametersStocasticPolicy}\log \policy{\action | \sstate, \modelParametersStocasticPolicy} \advantageFunction{\sstate, \action, \modelParametersValueFunction}
\end{equation}


%% file: tex/paper_MBAE.tex
\section{\MBAEText}
\label{sec:ModelBasedActionExploration}


In model-based \RL we are trying to solve the same policy parameter optimization as in \refequation{eqn:policyImprovement}.
To model the dynamics, we train one model to estimate the reward function and another to estimate the successor state.
The former is modeled as a direct prediction, while the 
latter is modeled as a distribution from which samples can be drawn via a GAN (generative adversarial network).

\subsection{Stochastic \MBAEText}

\begin{figure}[htb]
\includegraphics[width=0.95\linewidth]{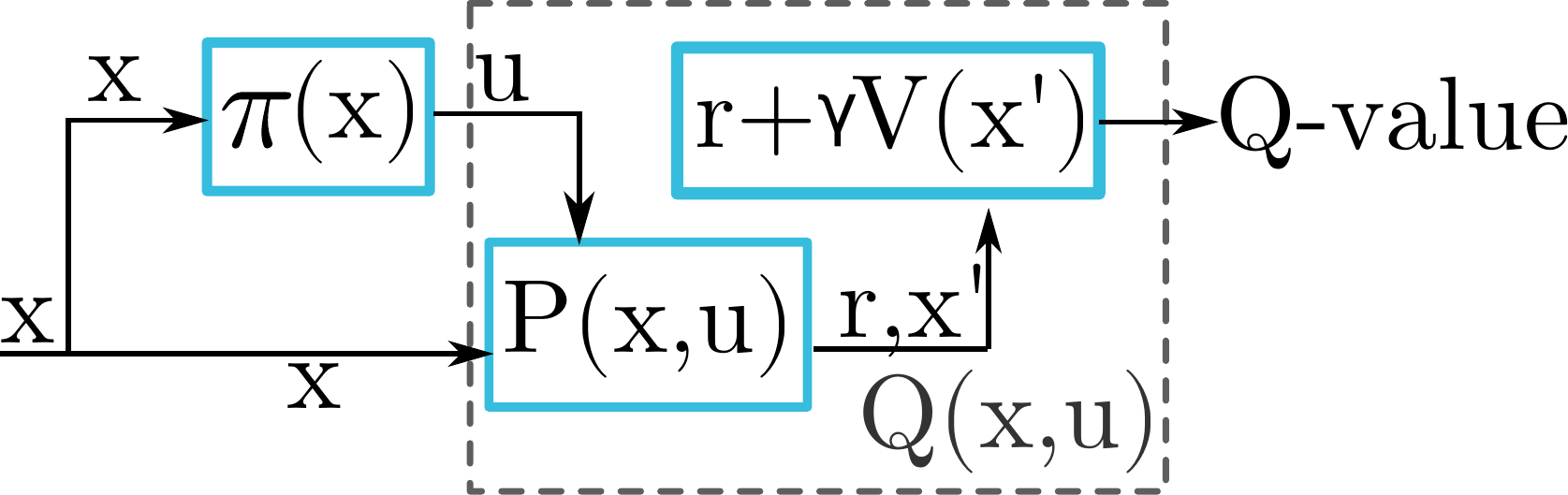}
\caption{Schematic of the \MBAEText design.  States $\observation$ are generated from the simulator,
  the policy produces an action $\action$, $\observation$ and $\action$ are used to predict the next
  state $\observation_{\ttime+1}$.  The gradient is computed back through the value function to give
  the gradient of the state $\grad\observation_{\ttime+1}$ that is then used to compute a gradient
  that changes the action by $\Delta\action$ to produce a predicted state with higher value.  }
\label{fig:mbae-model}
\end{figure}

A diagram of the \SMBAE method is shown in \refFigure{fig:mbae-model}.
With the combination of the \forwardDyanmicsText model and a value function, an action-valued function is constructed.
Using \SMBAE, action gradients can be computed and used for exploration.

By using a stochastic transition function the gradients computed by \SMBAE are non-deterministic.
\refAlgorithm{alg:compute-action_delta} shows the method used to compute action gradients when
predicted future states are sampled from a distribution.  We use a \GAN~\citep{NIPS2014_5423} to
model the stochastic distribution.  Our implementation closely
follows~\citep{DBLP:journals/corr/IsolaZZE16} that uses a \cGAN and combines a \MSE loss with the
normal \GAN loss.  We expect the simulation dynamics to have correlated terms, which the \GAN
can learn.
\begin{algorithm}[ht!]
\caption{Compute Action Gradient}
\label{alg:compute-action_delta}
\begin{algorithmic}[1]

\Function{getActionDelta}{$\sstate_{\ttime}$}
	\State{$\hat{\action}_{\ttime} \leftarrow \policy{\action | \observation_{\ttime}, \modelParametersStocasticPolicy}$}
	\State{$\noiseParam \leftarrow \normalDistribution{0}{1.0}$}
	\State{$\hat{\sstate}_{\ttime+1} \leftarrow \forwardDynamics{\observation_{\ttime+1} |\observation_{\ttime}, \hat{\action}_{\ttime}, \noiseParam, \modelParametersForwardDynamics}$}
	\State{$\grad\hat{\sstate}_{\ttime+1} \leftarrow \grad_{\sstate_{\ttime+1}} \valueFunction{\hat{\sstate}_{\ttime+1} | \modelParametersValueFunction}$}
	\State{$\hat{\action}_{\ttime} \leftarrow \hat{\action}_{\ttime} + \learningRate_{\action} \grad_{\hat{\action}_{\ttime}} \forwardDynamicsMEAN{\grad\hat{\sstate}_{\ttime+1} | \observation_{\ttime}, \hat{\action}_{\ttime}, \noiseParam, \modelParametersForwardDynamicsMEAN}$}
	\State \Return $\hat{\action}_{\ttime}$
\EndFunction

\end{algorithmic}
\end{algorithm}
$\learningRate_{\action}$ is a learning rate specific to \SMBAE and $\noiseParam$ is the random noise sample used by the \cGAN.
This exploration method can be easily incorporated into RL algorithms.
The pseudo code for using \SMBAE is given in \refAlgorithm{alg:SMBAE}. 

\begin{algorithm}[h!]
\caption{\SMBAE algorithm}
\label{alg:SMBAE}
\begin{algorithmic}[1]
	\State{Randomly initialize model parameters}
	\While{not done}
	
		\While{Simulate episode}
			\If{generate exploratory action}
				\State{$\action_{\ttime} \leftarrow \policy{\action | \observation_{\ttime}, \modelParametersStocasticPolicy}$}
				\If {$ \function{Uniform}(0,1) < p$}
					\State { $\action_{\ttime} \leftarrow \action_{\ttime} + $ $\function{GetActionDelta}(\observation_{\ttime})$}
				\EndIf
				\Else
					\State{$\action_t \leftarrow \policyMean{\observation_{\ttime} | \modelParametersPolicy}$}
			\EndIf
		\EndWhile
		\State{Sample batch $\{\experianceTuple_j = (\observation_j, \action_j, \reward_j, \observation'_j)\}$ from $\experianceMemory$}
		\State{Update value function, policy and \forwardDyanmicsText given $\{\experianceTuple_j\}$}
	\EndWhile
\end{algorithmic}
\end{algorithm}

\subsection{DYNA}

In practise the successor state distribution produced from \SMBAE will differ from the environment's
true distribution.  To compensate for this difference we perform additional training updates on the
value function, replacing the successive states in the batch with ones produced from
$\forwardDynamics{\cdot | \observation_{\ttime}, \action_{\ttime}, \noiseParam,
  \modelParametersForwardDynamics}$.  This helps the value function better estimate future
discounted reward for states produced by \MBAE.  This method is similar to
\DYNA~\citep{Sutton90integratedarchitectures,sutton1991dyna}, but here we are performing these
updates for the purposes of conditioning the value function on the transition dynamics model.

\section{Connections to Policy Gradient Methods}

Action-valued functions can be preferred because they model the effect of taking specific actions
and can also implicitly encode the policy.  However, performing a value iteration update over the
all actions
is intractable in continuous action spaces.

\begin{equation}
\label{eqn:action-value-function}
L(\theta) = \expectation[\reward + \discountFactor \max_{\action' \in \actionSpace} Q_{\pi(\observation_{\ttime+1})}(\observation_{\ttime+1}, \action', \theta) - Q_{\pi(\observation|\modelParametersStocasticPolicy)}(\observation, \action, \theta)]
\end{equation}
\DPG~\citep{silver2014deterministic} compensates for this issue by
linking the value and policy functions together allowing for gradients to be passed from the value
function through to the policy.  The policy parameters are then updated to increase the action-value
function returns.
This method has been successful~\citep{DBLP:journals/corr/HPHETS15} but has stability challenges~\citep{DBLP:journals/corr/HausknechtS15a}.

More recently \SVG~\cite{NIPS2015_5796} has been proposed as a method to unify model-free and
model-based methods for learning continuous action control policies.  The method learns a stochastic
policy, value function and stochastic model of the dynamics that are used to estimate policy
gradients.  While \SVG uses a similar model to compute gradients to optimize a policy, here we use
this model to generate more informed exploratory actions.

%% file: tex/paper_Results.tex
\section{Results}
\label{sec:results}

\MBAE is evaluated on a number of tasks, including: \Membrane robot simulation of
move-to-target and stacking, \Membrane robot hardware move-to-target, OpenAIGym \halCheetah,
OpenAIGym \reacher, \character simulation and N-dimensional particle navigation.  
The supplementary video provides a short overview of these systems and tasks.
The method is
evaluated using the \CACLA stochastic policy \RL learning algorithm~\citep{vanHasselt2012}.
\CACLA updates the policy mean using \MSE for actions that have positive advantage.

\subsection{N-Dimensional Particle}
\begin{figure}[h!]
\begin{centering}
\subcaptionbox{\label{fig:particle-world-2d-agent} Nav environment and current policy}{\includegraphics[width=0.92\columnwidth]{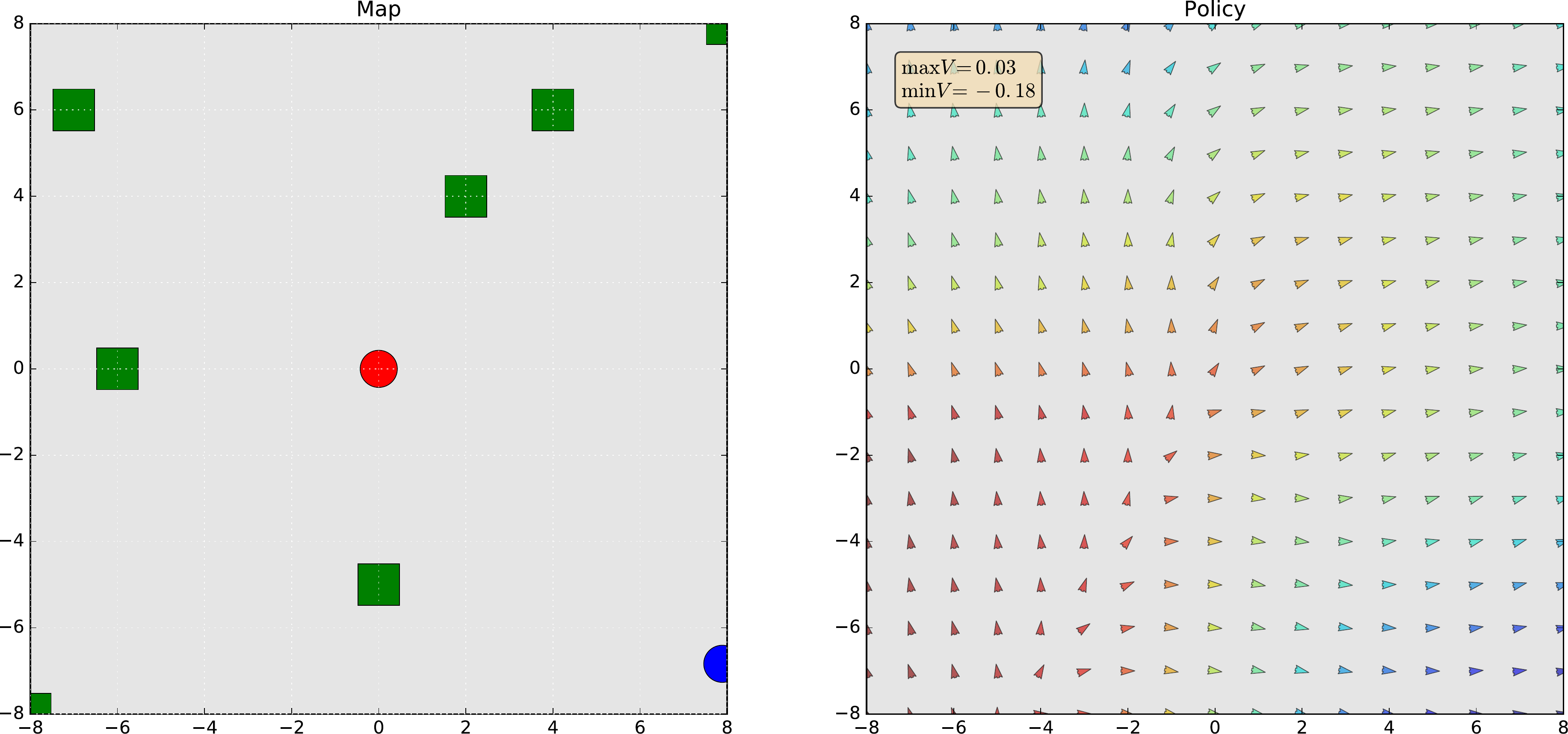}} \\
\subcaptionbox{\label{fig:particle-world-2d-fdError} \forwardDyanmicsText error}{\includegraphics[width=0.48\columnwidth]{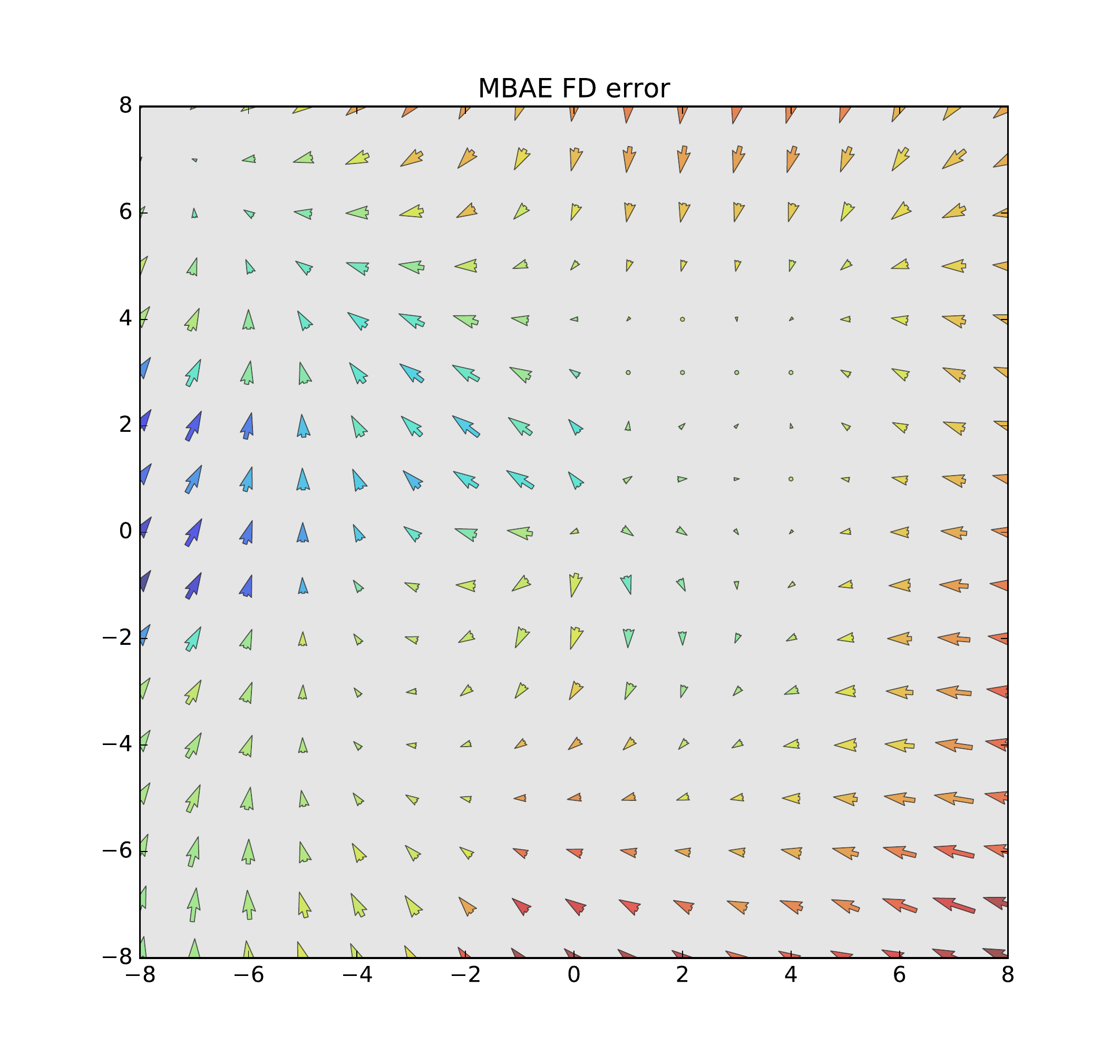}}
\subcaptionbox{\label{fig:particle-world-2d-mbae} \MBAE direction}{\includegraphics[width=0.48\columnwidth]{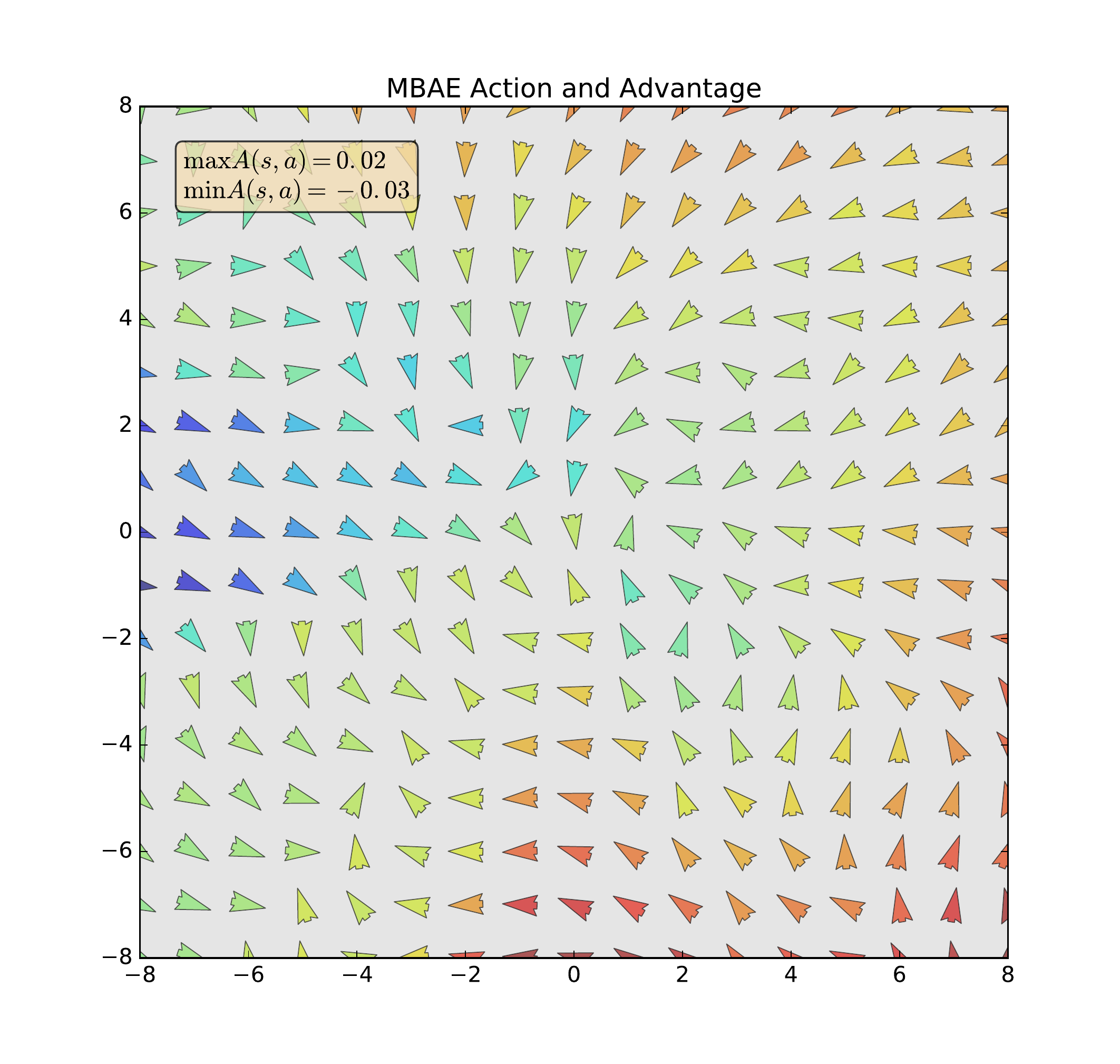}} \\
\caption{The figure (a) left is the current layout of the continuous grid world. The agent is blue
  and target location for the agent is the red dot and the green boxes are obstacles.  In (a) right,
  the current policy is shown as if the agent was located at each arrow 
  action to give the unit direction of the action.  The current value at each state is visualized by
  the colour of the arrows, red being the highest.  In (b) the error of the forward dynamics model
  is visualized as the distance between the successive state predicted and the actual successive
  states ($(\observation + \action) - \forwardDynamics{\observation, \action}$).  (c) is the unit
  length action gradient from \MBAE.  Only the first two dimensions of the state and action are
  visualized here.  }
\label{fig:grid-world-2d}
\end{centering}
\end{figure}

This environment is similar to a continuous action space version of the common grid world problem.
In the grid world problem the \emph{agent} (blue dot) is trying to reach a \emph{target location}
(red dot), shown in the left of \refFigure{fig:particle-world-2d-agent}.  In this version the agent
receives reward for moving closer to its goal ($ r = || agent_{pos} - target_{pos}||_{2}$).  This
problem is chosen because it can be extended to an N-dimensional world very easily, which is
helpful as a simple evaluation of scalability as the action-space dimensionality increases.
We use a 10D version here \citep{DBLP:journals/corr/TamarLA16,DBLP:journals/corr/FinnYFAL16}.

\refFigure{fig:grid-world-2d} shows a visualization of a number of components used in \SMBAE. 
In \refFigure{fig:mbae-compare-cacla} we compare the learning curves of using a
standard \CACLA learning algorithm and one augmented with \SMBAE for additional action
exploration.  The learning curves show a significant improvement in learning speed and policy
quality over the standard \CACLA algorithm.
We also evaluated the impact of pre-training the deterministic \forwardDyanmicsText model for \MBAE.
This pre-training did not provide noticeable improvements.

\subsection{2D Biped Imitation}

\begin{figure}[h!]
\centering
\begin{minipage}{.45\linewidth}
\subcaptionbox{\label{fig:envs-biped} \character }{\includegraphics[width=0.95\linewidth]{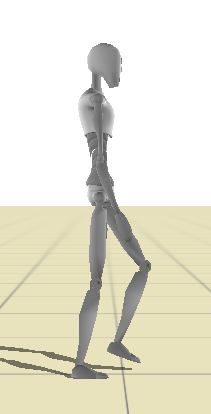}}
\end{minipage}
\begin{minipage}{.45\linewidth}
\subcaptionbox{\label{fig:envs-reacher} \reacher}{\includegraphics[width=0.95\linewidth]{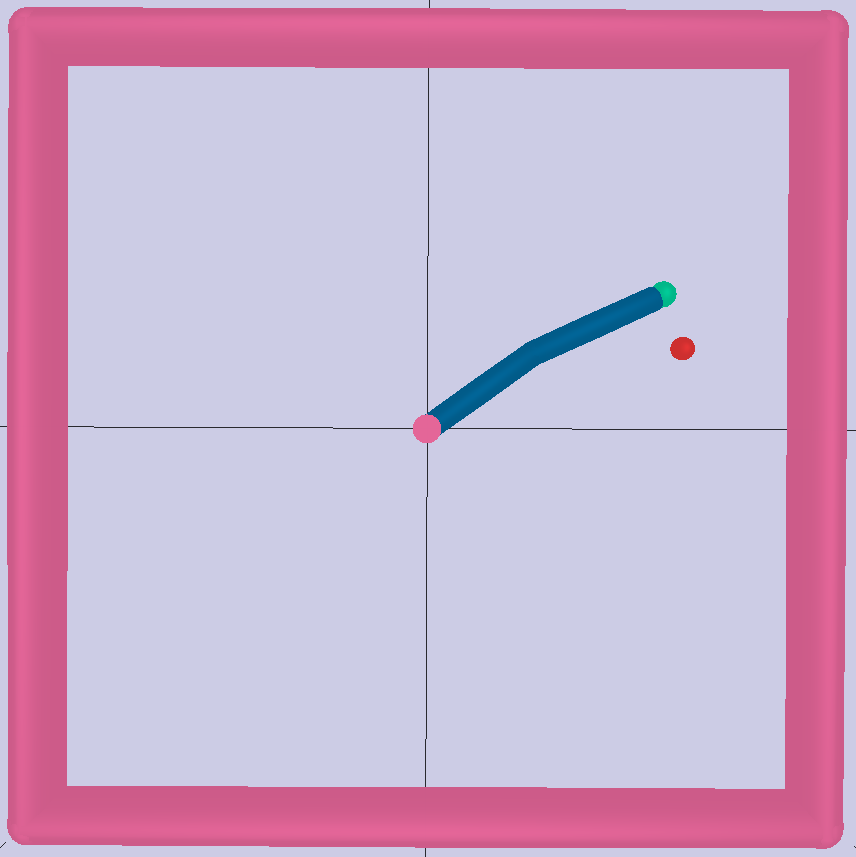}} 
\\
\subcaptionbox{\label{fig:envs-halfcheetah} \halCheetah}{\includegraphics[width=0.95\linewidth]{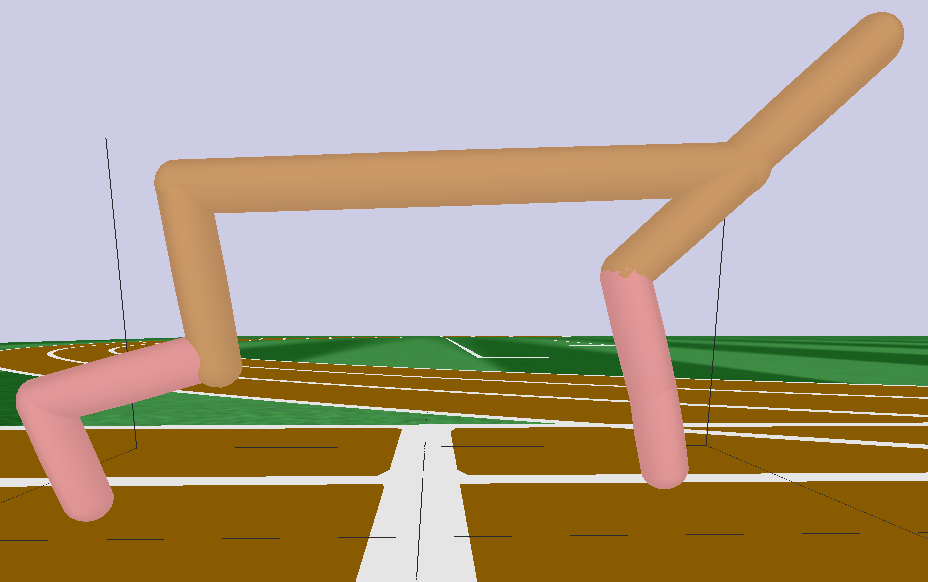}}
\end{minipage}
\caption{Additional environments \MBAE is evaluated on.}
\label{fig:envs}
\end{figure}

In this environment the \agent is rewarded for developing a 2D walking gait. 
Reward is given for matching an overall desired velocity and for matching a given reference motion.
This environment is similar to~\citep{Peng:2017:LLS:3099564.3099567}.
The \character used in the simulation is shown in~\refFigure{fig:envs-biped}.

In \refFigure{fig:LLC2D-MBAE-compare},  five evaluations are used for the \character and the mean learning curves are shown. 
In this case \MBAE consistently learns $5$ times faster than the standard \CACLA algorithm.
We further find that the use of \MBAE also leads to improved learning stability and more optimal policies.

\subsection{Gym and Membrane Robot Examples}

\begin{figure*}[h!]
\centering
\subcaptionbox{\label{fig:mbae-compare-cacla} Particle 10D}{\includegraphics[width=0.32\linewidth]{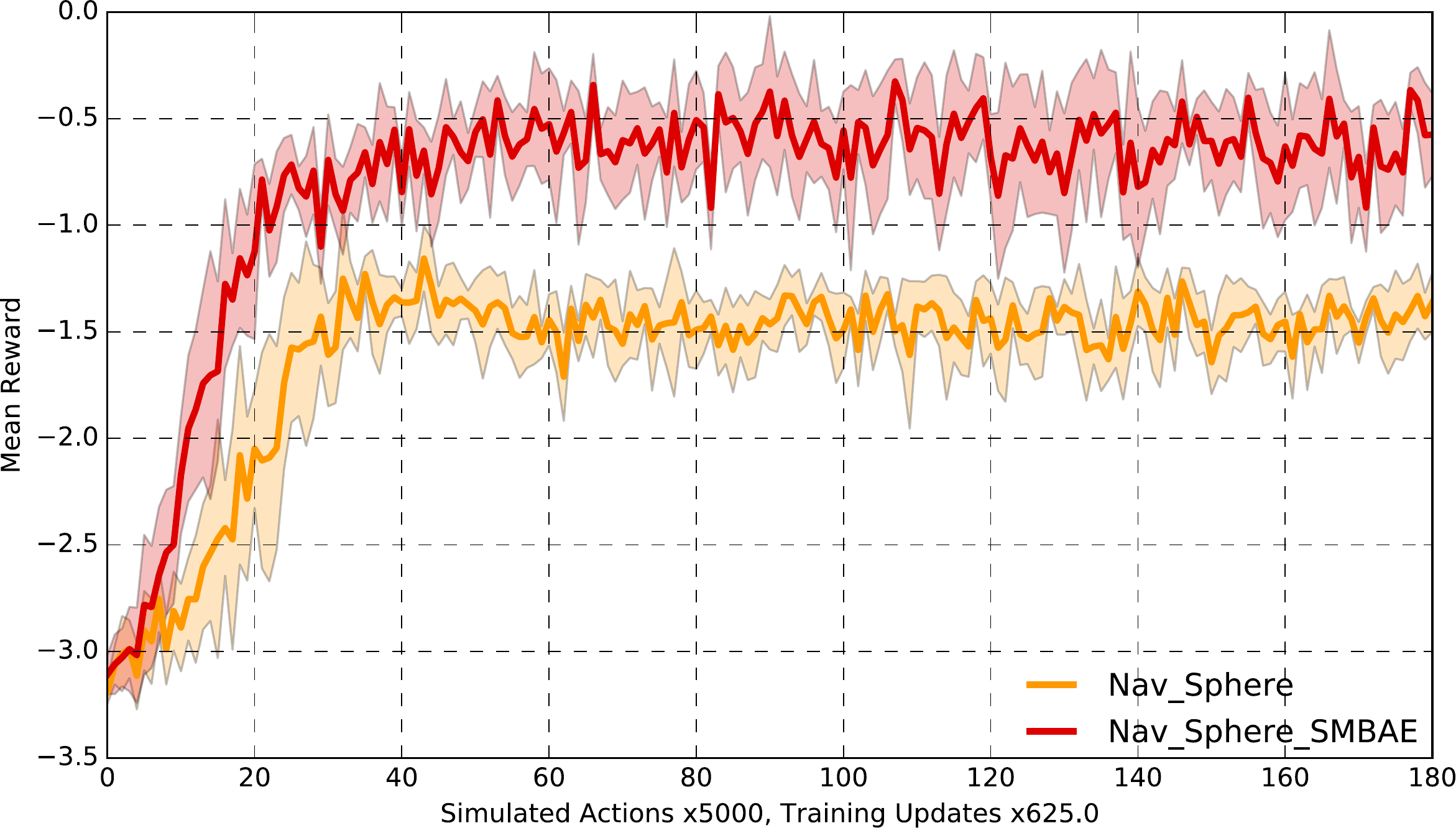}}
\subcaptionbox{\label{fig:LLC2D-MBAE-compare} 2D PD Biped}{\includegraphics[width=0.32\linewidth]{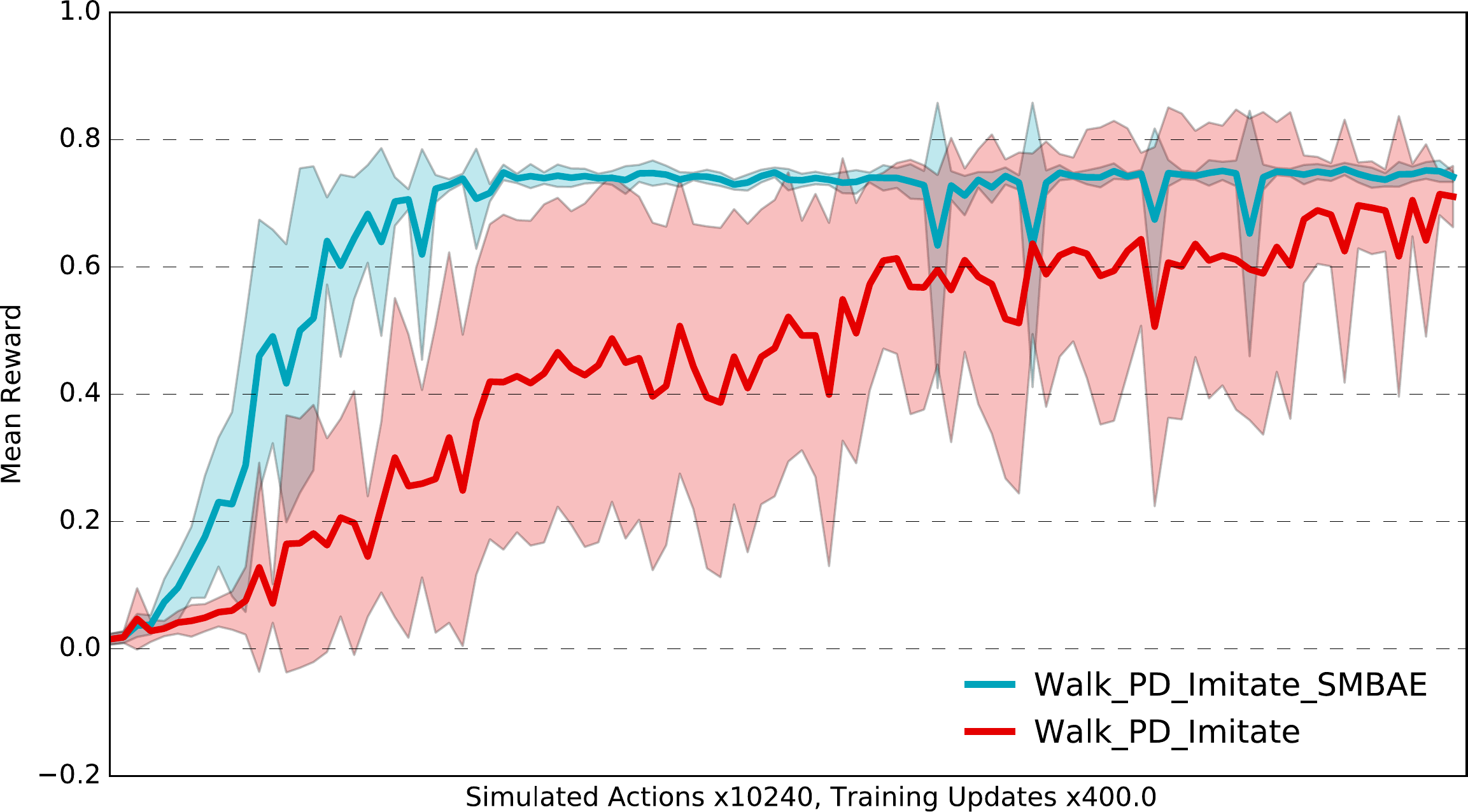}}
\subcaptionbox{\label{fig:robot-examples-reacher} Reacher 2D}{\includegraphics[width=0.32\linewidth]{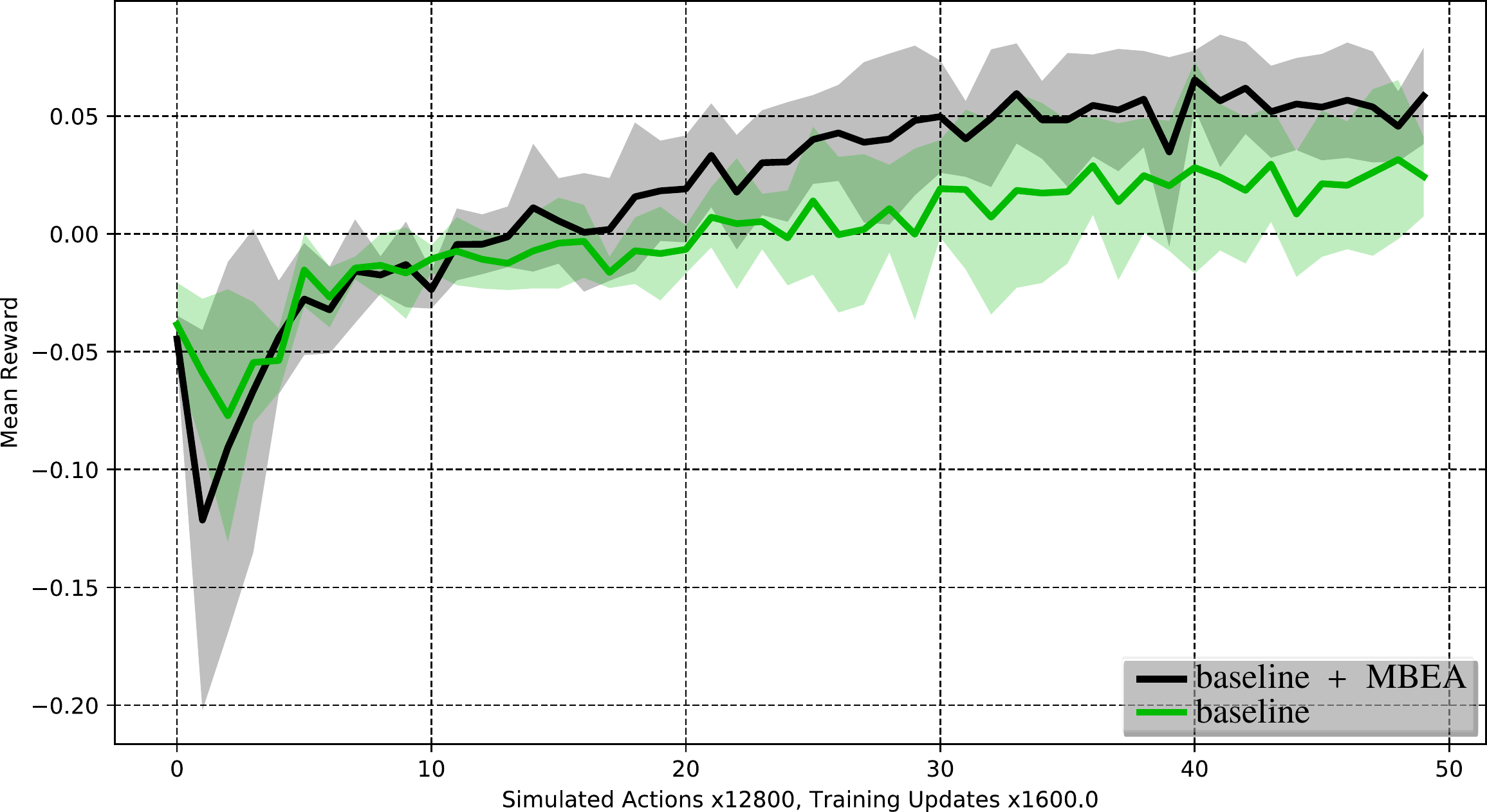}} \\
\subcaptionbox{\label{fig:robot-examples-half-cheetah} Half-Cheetah}{\includegraphics[width=0.32\linewidth]{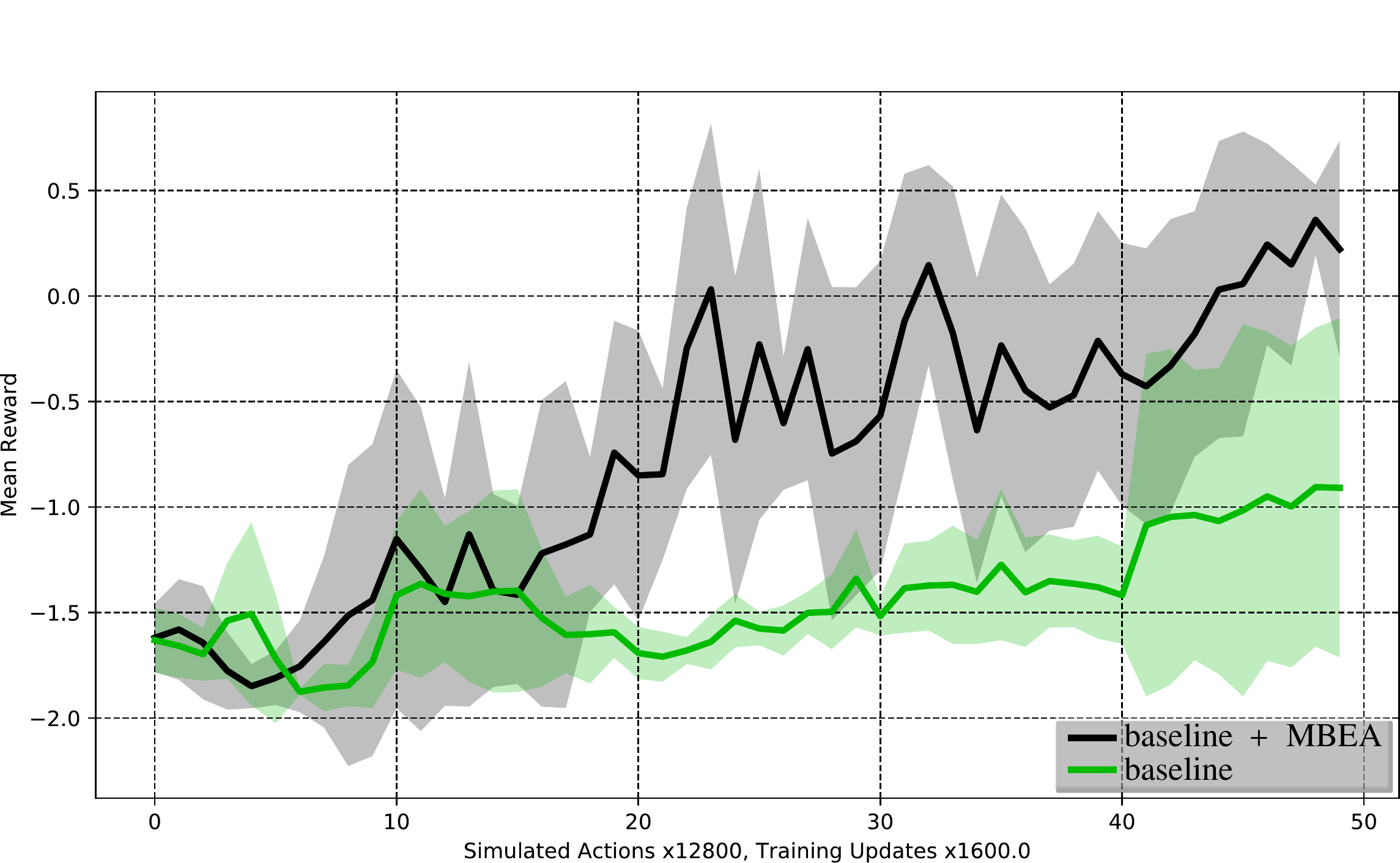}}
\subcaptionbox{\label{fig:robot-examples-membrane-movetotarget}Membrane Target}{\includegraphics[width=0.32\linewidth]{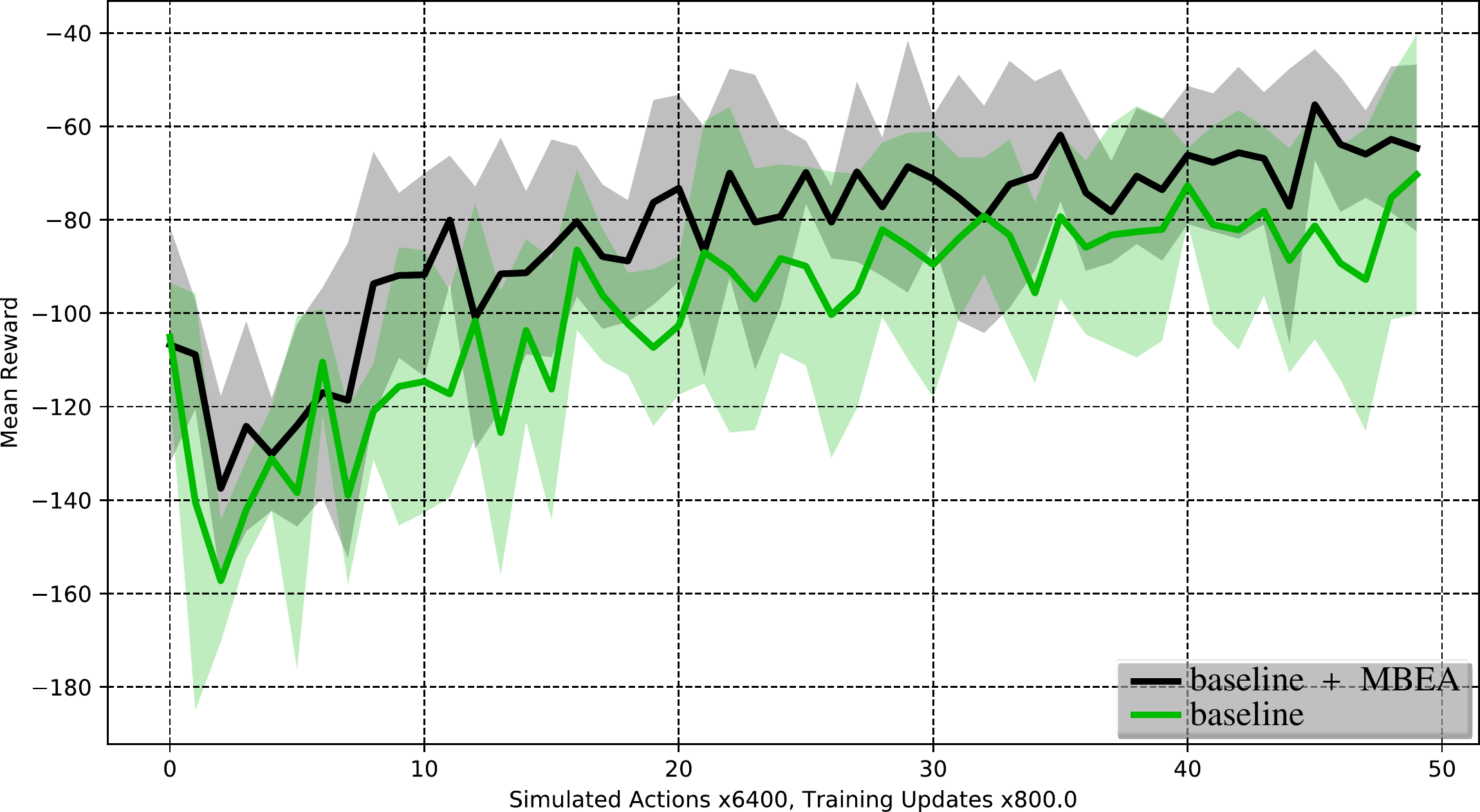}} 
\subcaptionbox{\label{fig:robot-examples-membrane-stack}Membrane Stack}{\includegraphics[width=0.32\linewidth]{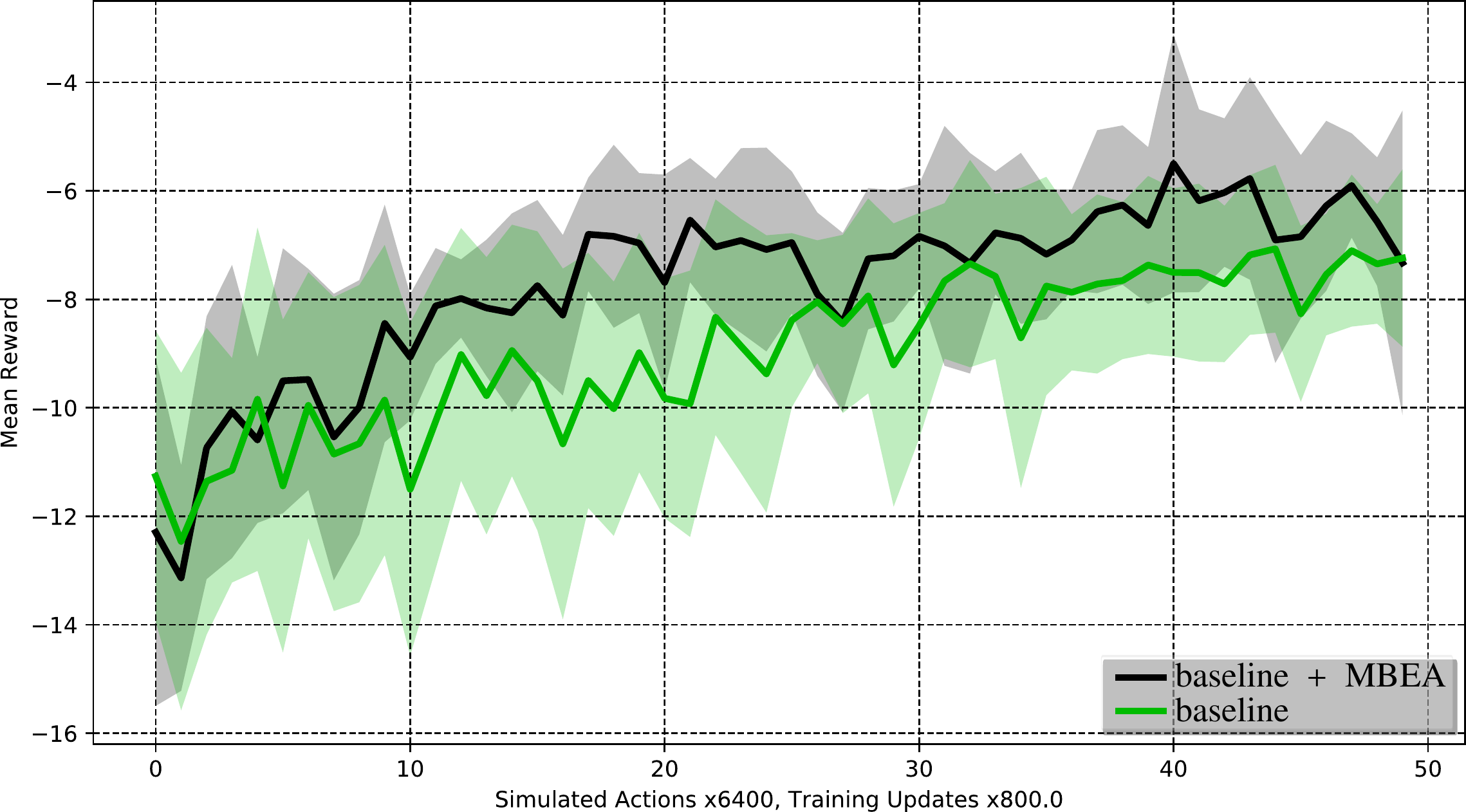}}
\caption{Comparisons of using the \CACLA learning method with and without \MBAE. These performance curves are the average of $5$ separate simulation with different random seeds.}
\label{fig:robot-examples}
\end{figure*}

We evaluate \MBAE on two environments from openAIGym, 
\reacher~\refFigure{fig:envs-reacher} and \halCheetah~\refFigure{fig:envs-halfcheetah}. 
\MBAE does not significantly improve the learning speed for the \reacher. 
However, it results in a higher value policy~\refFigure{fig:robot-examples-reacher}.
For the \halCheetah \MBAE provides a significant learning improvement~\refFigure{fig:robot-examples-half-cheetah}, 
resulting in a final policy with more than $3$ times the average reward.  

\begin{figure}[h!]
\centering
\subcaptionbox{\label{fig:robot-examples-membrane-hardware-mbae}Membrane Target}{\includegraphics[width=0.82\linewidth]{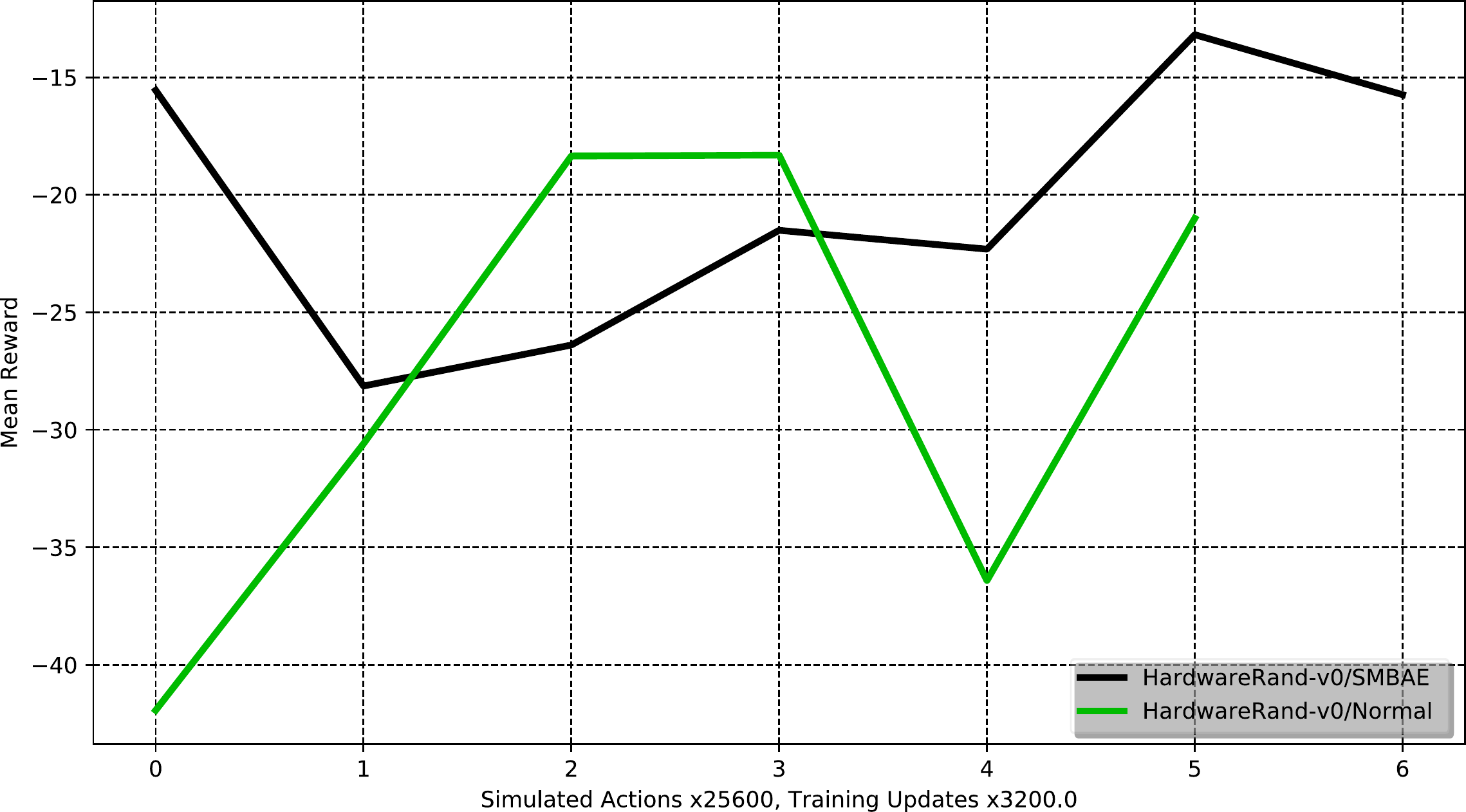}} 
\subcaptionbox{\label{fig:robot-examples-membrane-hardware-normal}Membrane Camera}{\includegraphics[width=0.42\linewidth]{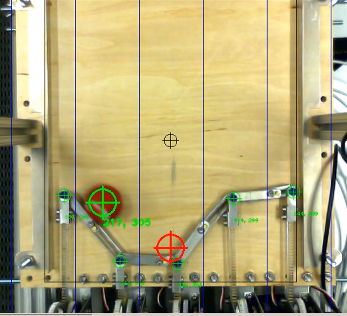}}
\subcaptionbox{\label{fig:robot-examples-membrane-software-stack}Membrane Stack}{\includegraphics[width=0.42\linewidth]{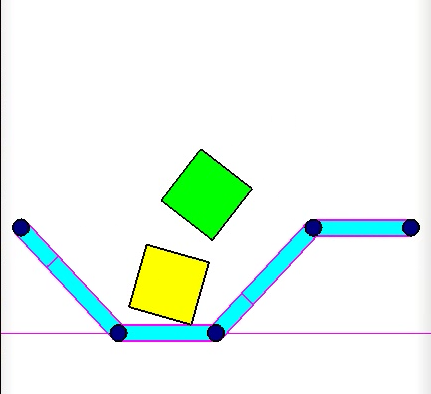}}

\caption{(a) Comparison of using the \MBAE on the physical robot task. (b) is the camera view the robot uses to track its state and (c) is a still frame from the simulated box \textit{stacking} tasks.}
\label{fig:robot-examples-membrane-hardware}
\end{figure}

Finally, we evaluate \MBAE on a simulation of the \textit{juggling} \Membrane robot shown
in~\refFigure{fig:Membrane-hardware}.  The under-actuated system with complex dynamics and regular
discontinuities due to contacts make this a challenging problem.  The results for two tasks
that include attempting to stack one box on top of another and a second task to move a ball to a
target location are shown in~\refFigure{fig:robot-examples-membrane-stack} and
~\refFigure{fig:robot-examples-membrane-movetotarget}.  For both these environments the addition of
\MBAE provides only slight improvements.  We believe that due to the complexity of this learning task,
it is difficult to learn a good policy for this problem in general.  The simulated version of the
\textit{membrane-stack} task is shown in~\refFigure{fig:robot-examples-membrane-software-stack}.

We also asses \MBAE on the \Membrane robot shown in~\refFigure{fig:Membrane-hardware}.  OpenCV is
used to track the location of a ball that is affected by the actuation of $5$ servos that
cause $5$ pins to move linearly, shown in~\refFigure{fig:robot-examples-membrane-hardware-normal}.
The $5$ pins are connected by passive prismatic joints that form the \textit{membrane}.  The robot
begins each new episode by resetting itself which involves tossing the ball up and randomly
repositioning the membrane. Please see the accompanying video for details.  We transfer the
\textit{movetotarget} policy trained in simulation for use with the \Membrane robot.  We show the
results of training on the robot with and without \MBAE for $\sim3$ hours each
in~\refFigure{fig:robot-examples-membrane-hardware-mbae}.  Our main objective here is to demonstrate
the feasibility of learning on the robot hardware; our current results are only 
from a single training run for each case. 
With this caveat in mind, \MBAE appears to support improved learning.
We believe that this is related to the \forwardDyanmicsText model adjusting to the new state distribution of the
robot quickly.  

\subsection{Transition Probability Network Design}

We have experimented with many network designs for the \forwardDyanmicsText model.
We have found that using a DenseNet~\citep{Huang_2017_CVPR} works well and increases the models accuracy.
We use dropout on the input and output layers, as well as the inner layers, to reduce overfitting. 
This makes the gradients passed through the \forwardDyanmicsText model less biased.

%% file: tex/paper_Discussion.tex
\section{Discussion}
\label{sec:discussion}

\paragraph{Exploration Action Randomization and Scaling}
Initially, when learning begins, the estimated policy gradient is flat, making \MBAE actions $ \sim
0$.  As learning progresses the estimated policy gradient gets sharper leading to actions produced
from \MBAE with magnitude $>> 1$.  By using a normalized version of the action gradient, we maintain
a reasonably sized explorative action, this is similar to the many methods used to normalize
gradients between layers for deep learning~\citep{2016arXiv160706450L,DBLP:journals/corr/IoffeS15}.
However, with normalized actions, we run the risk of being overly deterministic in action
exploration.  The addition of positive Gaussian noise to the normalized action length
helps compensate for this.  Modeling the transition dynamics stochasticity allows us to generate
future states from a distribution, further increasing the stochastic nature of the action
exploration.

\paragraph{\forwardDyanmicsText Model Accuracy}
Initially, the models do not need to be significantly accurate. They only have to perform better than random (Gaussian) sampling.
We found it important to train the \forwardDyanmicsText model while learning. 
This allows the model to adjust and be most accurate for the changing state distribution observed during training.
This makes it more accurate as the policy converges.

\paragraph{\MBAE Hyper Parameters}
To estimate the policy gradient well and to maintain reasonably accurate value estimates, Gaussian
exploration should still be performed.  This helps the value function get a better estimate of the
current policy performance.  From empirical analysis, we have found that sampling actions from \MBAE
with a probability of $0.25$ has worked well across multiple environments.  The learning progress
can be more sensitive to the action learning rate $\learningRate_{\action}$.  We found that
annealing values between $1.0$ and $0.1$ \MBAE assisted learning.  The form of normalization that
worked the best for \MBAE was a form of batchnorm, were we normalize the action standard deviation
to be similar to the policy distribution.

One concern could be that \MBAE is benefiting mostly from the extra training that is being seen for the value function.
We performed an evaluation of this effect by training \MBAE without the use of exploratory actions from \MBAE. 
We found no noticeable impact on the learning speed or final policy quality.

\subsection{Future Work}

It might still be possible to further improve \MBAE by pre-training the \forwardDyanmicsText model
offline.  As well, learning a more complex \forwardDyanmicsText model similar to what has been done
in~\citep{DBLP:journals/corr/MishraAM17} could improve the accuracy of the \MBAE generated actions.
It might also be helpful to learn a better model of the reward function using a method similar
to~\citep{Silver2016}.  One challenge is the addition of another \emph{step size} $\learningRate$
for how much action gradient should be applied to the policy action, and it can be non-trivial to
select this step size.

While we believe that the \MBAE is promising, the learning method can suffer from stability issues when 
the value function is inaccurate, leading to poor gradients.  We are
currently investigating methods to limit the KL divergence of the policy between updates.
These constraints are gaining popularity in recent RL methods~\citep{TRPO}.  This should reduce the
amount the policy shifts from parameter updates, further increasing the stability of learning.  The
\Membrane related tasks are particularly difficult to do well on; even after significant training
the policies could still be improved.  Lastly, while our focus has been on evaluating the method on many
environments, we would also like to evaluate \MBAE in the context of additional \RL algorithms, such as PPO or Q-Prop, to
further assess its benefit.





%% file: tex/paper_appendix.tex
\addcontentsline{toc}{chapter}{Appendix}
\section{Appendix}
\label{sec:Appendix}

\subsection{Max Over All Actions, Value Iteration}

By using \SMBAE in an iterative manner, for a single state ($\myState_{\ttime}$), it is possible to compute the max over all actions. This is a form of value iteration over the space of possible actions. It has been shown that embedding value iteration in the model design can be very beneficial~\citep{DBLP:journals/corr/TamarLA16}
 The algorithm to perform this computation is given in~\refAlgorithm{alg:Advantage-optimization}.

\begin{algorithm}[ht!]
\caption{Action optimization}
\label{alg:Advantage-optimization}
\begin{algorithmic}[1]

	\State{$\hat{\action}_{\ttime} \leftarrow \policy{\observation_{\ttime}| \modelParametersPolicy}$}
\While{not done}
	\State{ $\hat{\action}_{\ttime} \leftarrow \hat{\action}_{\ttime} + \function{GetActionDelta}(\hat{\action}_{\ttime})$}
\EndWhile
\end{algorithmic}
\end{algorithm}

\subsection{More Results}

\begin{figure}[h!]
\begin{centering}
\subcaptionbox{\label{fig:ppo-results-particleGame} \PPO game}{\includegraphics[width=0.48\columnwidth]{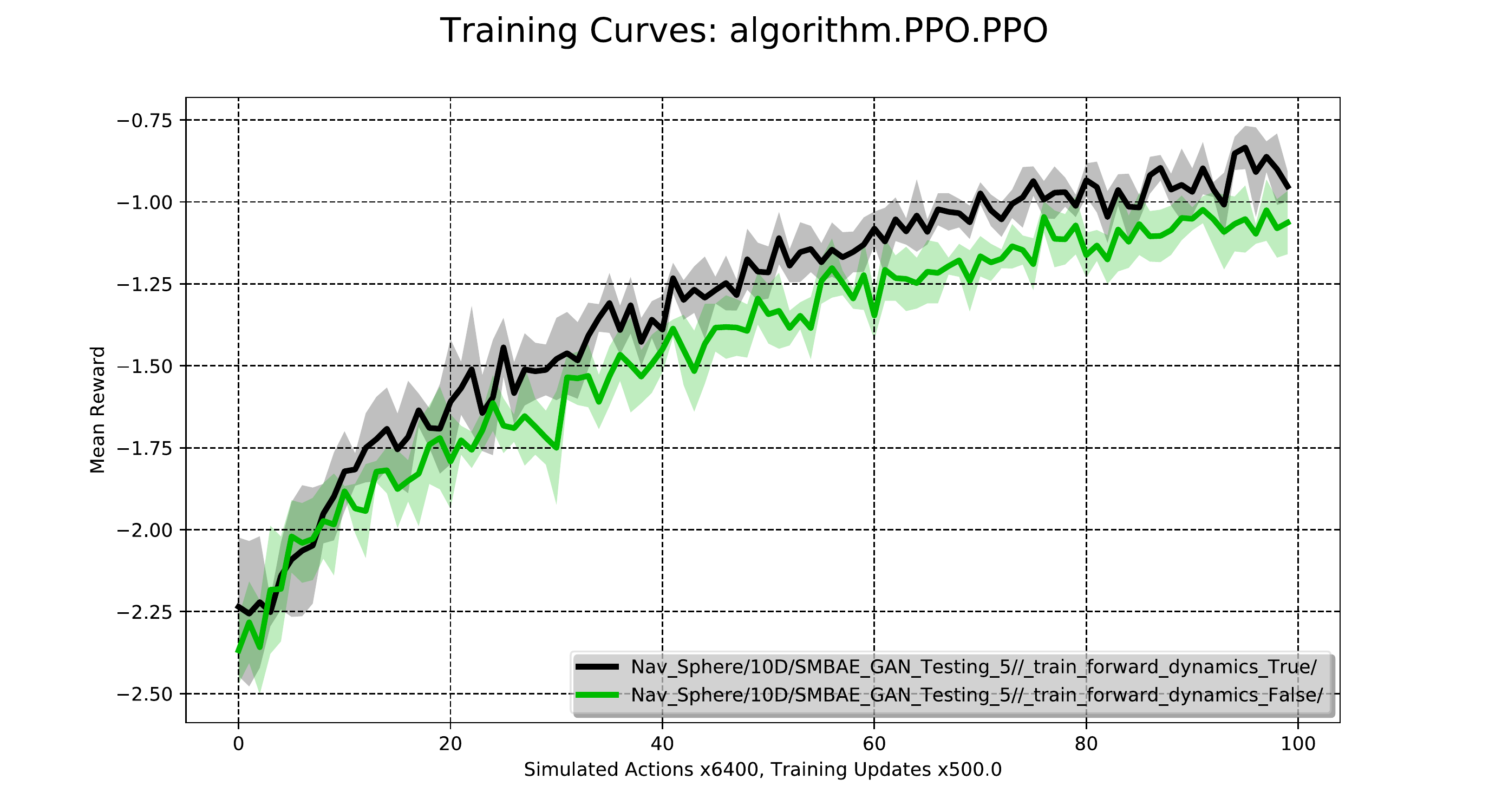}} 
\subcaptionbox{\label{fig:cacla-on-policy-results-particleGame} on-policy \CACLA game}{\includegraphics[width=0.48\columnwidth]{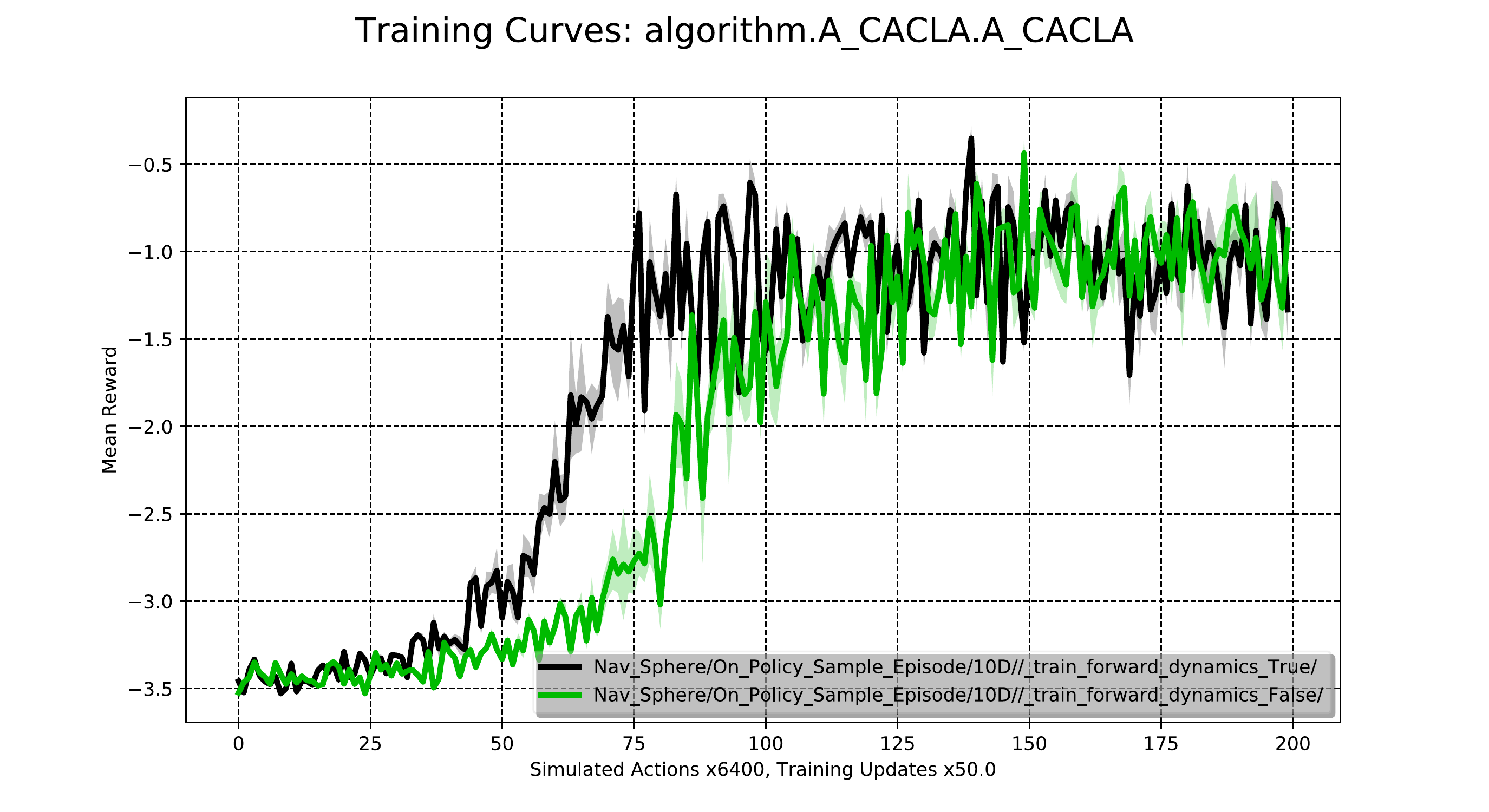}} \\
\caption{ (a) Result of Applying \MBAE to \PPO. In (b) we show that an on-policy version of \CACLA + \MBAE can learn faster than \CACLA alone. 
}
\label{fig:ppo-results}
\end{centering}
\end{figure}

We perform additional evaluation on \MBAE. 
First we use \MBAE with the \PPO~\cite{2017arXiv170706347S} algorithm in~\refFigure{fig:ppo-results-particleGame} to show that the method works with other learning algorithms.
We also created a modified version of \CACLA that is on-policy to further study the advantage of using \MBAE in this setting~\refFigure{fig:cacla-on-policy-results-particleGame}.
